\documentclass[11pt]{article}
\usepackage{pepadun}
\usepackage{float}

\title{Ensemble Deep Learning Approaches for AI-Altered Video Detection}
\author{%
    \begin{tabular}{c}
        \textsuperscript{1}Khan, Laiba \\
        \textsuperscript{2}Wu, Hung-Mao \\
        \textsuperscript{3}Lin, Wei \\
        \textsuperscript{4}Bi, Frank \\
        \textsuperscript{5}Abdelhadi, Yousef and \\
        \textsuperscript{6}Jung, Joshua \\[6pt]
        \textsuperscript{1,2,3,4,6}Department of Mathematical and Computational Sciences, University of Toronto Mississauga, \\
        \textsuperscript{5}Department of Mathematical and Computational Sciences, University of Toronto, \\ 
        University of Toronto, Mississauga, Canada \\[6pt]
        e-mail: \textsuperscript{1} llaiba.khan@mail.utoronto.ca, \\
        \textsuperscript{2} \href{mailto:hongmao.wu@mail.utoronto.ca}{\uline{hongmao.wu@mail.utoronto.ca}} \\
        \textsuperscript{3} \href{mailto:marcow.lin@mail.utoronto.ca}{\uline{marcow.lin@mail.utoronto.ca}}\\
        \textsuperscript{4} \href{mailto:frank.bi@mail.utoronto.ca}{\uline{frank.bi@mail.utoronto.ca}}\\
        \textsuperscript{5} \href{mailto:yousef.abdelhadi@utoronto.ca}{\uline{yousef.abdelhadi@utoronto.ca}} \\
        \textsuperscript{6} \href{mailto:josh.jung@utoronto.ca}{\uline{josh.jung@utoronto.ca}}
    \end{tabular}%
}
\date{}

\begin{document}

\maketitle

\begin{abstract}
The increasing accessibility of artificial intelligence has led to a rapid rise in AI-generated videos, making it more difficult to distinguish between real and manipulated content. Many existing detection methods rely on a single model and often struggle to generalize across different types of deepfakes. In this work, we developed a multimodal deepfake detection system that combines both audio and visual analysis using an ensemble of models. The system includes AASIST for audio-based detection, and EfficientNet, XceptionNet, and MesoNet for analyzing visual features in video frames. The pipeline takes a video as input, separates the audio, and extracts face frames using MTCNN. Each model produces a score indicating the likelihood of the input being fake. These scores are then combined using ensemble strategies, including mean averaging and stacking. Mean fusion provides a simple and stable baseline, while stacking uses a trained meta-model to learn how to combine predictions more effectively. Results show that while individual models perform well on the datasets they were trained on, their performance drops when tested on more diverse datasets. The ensemble approach helps improve overall robustness by combining predictions from multiple models, leading to more consistent performance across different types of deepfakes. This suggests that using both audio and visual information together is a more reliable approach for deepfake detection. Our results highlight generalization to unseen manipulations as the central open challenge, with average accuracy around 70\%.
\end{abstract}

\keywords{Convolutional Neural Networks; Deepfakes; Ensemble Learning; MTCNN; Video Authentication}

\section{INTRODUCTION}

In recent years, the rapid development of artificial intelligence (AI) has led to a large number of AI-generated videos appearing in people's lives. Free, widely-available AI tools have allowed anyone to create any kind of AI video, which not only takes a short amount of time but also requires no more than consumer-grade computers. As such, AI-generated videos have become incredibly prevalent throughout the internet, particularly on social media platforms. Although many AI videos have been generated for humorous purposes, other videos have hidden or harmful motivations, aiming to manipulate and direct public opinion \cite{narayan2022desi}\cite{corsi2024synthetic}. For example, a video-generation technique has been to swap the faces of individuals between videos, which were then used to impersonate, sexualize, or harass the individual. Such maliciously altered media have become known as deepfakes, a term first appearing in 2017 that has become widespread alongside its video counterparts in the past decade\cite{regan2025deepfakehistory}. From 2019 to 2023, the number of deepfake videos has increased by 650\% \cite{to2024analysis}, and it has become a pressing issue to know which videos were AI-generated.

Despite the ongoing efforts, the ever-evolving AI-generation techniques still outpace the current detection capabilities. The lack of such software allows malicious users to spread misinformation at an incredible speed, resulting in a trust crisis for social media users \cite{alkhazraji2023impact}.

This issue has inspired us to build an ensemble of AI models, each specialized in performing its own analysis, that is capable of producing an estimate by considering each individual model's prediction. With the ensemble, the ultimate goal is to investigate the limits and potential of universal AI video detection; through taking a weighted average of predictions from individual convolutional neural network (CNN) models and training with deepfake datasets to make a final prediction about a video. 

In addition, the ensemble can serve as a way to identify AI-generated content. When a social media user is suspicious that a certain video has been generated by AI, the user can download the video and upload it to the ensemble. The ensemble will return an overall confidence score, with additional scores from each individual model. The confidence score should clearly indicate to the user whether the video uploaded is a deepfake video. 

Although downloading every video that a user sees on social media to make a judgment is infeasible, we wish to contribute to the ongoing development of the deepfake detection technique. 
Our contributions are:
\begin{itemize}
    \item We present a multimodal ensemble combining audio (AASIST) and visual (EfficientNet, XceptionNet, MesoNet) models for deepfake detection, and compare five fusion strategies.
    \item We show that audio anti-spoofing models generalize poorly to in-the-wild multimodal data, while video models remain comparatively robust.
    \item We analyze how ensemble performance varies across manipulation types, showing that the system relies heavily on the visual modality while the audio modality contributes little.
\end{itemize}

\section{BACKGROUND AND LITERATURE REVIEW}
\subsection{Deepfake}
With its widespread development, AI has become an astonishingly convenient and easily accessible tool to create and spread false information. Shortly after the invention of generative adversarial networks (GANs) in 2016, the idea of deepfakes started to become well-known \cite{kismawadi2025impact}. The term deepfake combines two terms, "deep learning" and "fake". In most scenarios, deepfake refers to synthesized or manipulated voices, videos, or images generated by AI \cite{altuncu2024deepfake, brady2020deepfakes}.
\subsection{GAN}
A GAN consists of two deep neural networks; one is called the generator, and the other is called the discriminator. The generator is a generative model that tries to learn the data distribution behind a given dataset and is able to generate new data points based on the distribution. The discriminator's job is to discriminate real data from generated data\cite{wang2017generative}. During the generator training phase, the generator generates output and feeds it to the discriminator while the discriminator's weights are locked. The result from the discriminator is then fed into the generator to improve the generator. During the discriminator training phase, the generator's weights are locked, and the generated data is combined with real data to train the discriminator\cite{aggarwal2021generative}. This is inspired by a two-player zero-sum game in which both players try to maximize their scores. In real life, GANs are not only used to generate deepfake videos, but they are also being used in speech processing and detecting malware\cite{wang2017generative}.
\subsection{From GAN to Diffusion Model}
As technology has developed, diffusion models have gradually replaced GANs to become the primary technique in generating deepfake videos. A diffusion model incrementally adds Gaussian noise to an input image. At each stage, the noisy image is shown to a neural network. By doing this, the neural network is forced to learn what real data looks like and can identify the noise in the image. Once the neural network learns the noise and the real object, the network can subtract what it believes is noise from any image to create any real object\cite{kumar2025diffusion, croitoru2023diffusion, codingtech2025diffusion}. The diffusion model has several advantages compared to GANs\cite{liu2025review}:
\begin{itemize}
    \item GANs suffer from mode collapse, where the generative model generates the same picture over and over again because it finds a flaw in the discriminator
    \item GANs have to compress the input image, so some information is lost
    \item Diffusion models provide researchers with greater flexibility in designing tests and control over model behavior.
\end{itemize}
\subsection{Ensemble Method}
Ensemble methods refer to various strategies that use multiple models in order to achieve a final aggregated prediction with higher performance than each individual model \cite{ren2016ensemble}. In addition, ensemble methods can be applied for both classification and regression. There are many different ensemble methods that have been proven to be useful in different scenarios. However, the key to having a successful ensemble is to create diversity, which can be broken down into three categories:
\begin{itemize}
    \item data diversity: Generate multiple sub-datasets from the original dataset, and train an individual model on each sub-dataset (e.g., bagging)
    \item parameter diversity: Use different initial parameters and train on the same dataset
    \item structural diversity: Combine results from models with completely different architectures and underlying algorithms
\end{itemize}
Ensemble methods such as bagging and AdaBoost fit into data diversity. In contrast, other methods like divide-and-conquer leverage multiple forms of diversity\cite{ren2016ensemble}. In this work, we experiment with stacking, weighted voting, weighted average, mean, and majority voting.
\subsection{Stacked Generalization}
Previous studies have shown that stacking can perform at least as well as the best individual model in the ensemble.
A stacking ensemble consists of at least two base models and one meta-model. Stacking follows the following steps\cite{naimi2018stacked}:
\begin{enumerate}
    \item Split the dataset into N folds
    \item For each fold, train all base models on all other folds, and use that specific fold as the validation set. Repeat step 2 until predictions for every fold are produced.
    \item Create a meta-learner that analyzes each prediction with its ground truth, decides which base models make the most correct predictions, and then assigns weights accordingly
    \item Retrain all base models on all folds
    \item When making a prediction for a new data point, multiply each model's weight by its prediction, and aggregate all predictions to produce the final prediction
\end{enumerate}

The use of cross-validation ensures that the meta-model is trained on predictions from data not seen during the training of the base models, which helps reduce overfitting \cite{geron2019hands}. However, in practice, a separate validation (hold-out) set can be used instead of k-fold cross-validation to generate the meta-model inputs, as long as the predictions are obtained from data not seen during base model training \cite{brownlee2021stacking}. This trades off some robustness for reduced complexity which is what was implemented in our model.
\subsection{EfficientNet Architecture}
EfficientNet employs a compound scaling method that simultaneously scales network depth, width, and resolution using a fixed set of scaling coefficients. Unlike traditional approaches that scale these dimensions independently, this method uses a single compound coefficient ($\phi$) to ensure a balanced expansion, optimizing both accuracy and computational efficiency \cite{tan2019efficientnet}. The EfficientNet has 8 versions, ranging from B0 to B7. Notably, each version consists of multiple layers of Mobile Inverted Bottleneck Convolution (MBConv). Each layer uses a squeeze-and-excitation module to allow EfficientNet to focus on the most important channel in an image. Besides these, EfficientNet uses a swish activation function and common dropout method with stochastic regularization method to reduce overfitting. 
    
While the general-purpose EfficientNet originally output 1000 classes, the deepfake EfficientNet adds a final classification head that maps the extracted features to a binary output\cite{keras_efficientnet}.
\subsection{XceptionNet Architecture}

XceptionNet is a CNN based on depthwise separable convolutions proposed by Chollet \cite{chollet2017xception}. It consists of 36 depthwise separable convolution layers, divided into 14 modules, which are connected using linear residual connections except for the first and the last module. In the XceptionNet architecture, data first goes through the entry flow, then through a middle flow, which is repeated 8 times, and finally through an exit flow. Due to the use of depthwise separable convolutions, it can efficiently extract features from image patches and encode them into a linear feature vector, where each feature vector represents the potential manipulation artifacts in the image patch \cite{shah2022xceptionvit}. The resulting linear feature vector can be used by subsequent classification layers to classify the image as real or fake. XceptionNet has also demonstrated its performance in a competition known as the Deepfake Detection Challenge (DFDC) held by Kaggle, where the top-performing solutions had either used XceptionNet or EfficientNet-B7.

The original XceptionNet architecture was designed for the ImageNet dataset, which uses a 1000-class single-label classification task. 
For the purpose of deepfake detection, we modified the final classification layer to perform binary classification. In addition, we applied a dropout rate of 0.5 to reduce overfitting, following the original XceptionNet design \cite{chollet2017xception}. 
\subsection{MesoNet Architecture}

The MesoNet architecture remained largely unchanged compared to its original implementation\cite{afchar2018mesonet}, with the Meso4 model architecture selected to train and make predictions. The Meso4 architecture has two main sections: first, 4 consecutive layers each performing convolutions, ReLU activation function, and max pooling; second, 2 fully-connected dropout layers. After the two main sections, a sigmoid activation function is used to produce the final classification.

\subsection{AASIST Architecture}
The AASIST (Audio Anti-Spoofing using Integrated Spectro-Temporal Graph Attention Networks) model is designed to analyze the audio from each video and classify it as either bonafide (real) or spoofed \cite{aasist2022}. Unlike standard convolutional neural networks (CNNs) that operate on image data, AASIST processes raw audio waveforms and focuses on learning both temporal and spectral characteristics of the signal.

The architecture begins with convolutional layers that extract low-level features from the input waveform. These features include short-term frequency patterns and local temporal variations, which are important for identifying artifacts introduced by text-to-speech and voice conversion methods \cite{tak2021rawgat}. Similar to CNNs in image processing, these layers act as feature extractors by capturing localized patterns in the signal.

Following the convolutional layers, AASIST incorporates spectro-temporal graph attention layers. These layers model the relationships between different time-frequency regions by representing them as nodes in a graph and learning attention weights between them. This allows the model to capture long-range dependencies and interactions across the audio signal, which are difficult to detect using standard convolutional operations alone. As a result, the model is able to identify more complex and distributed spoofing artifacts.

The final stage of the architecture consists of a classification head that outputs logits for two classes, bonafide and spoof. These logits are passed through a softmax function to produce a probability distribution, where the spoof probability is used as the final prediction score.


\section{RESEARCH METHODOLOGY}

All four of the aforementioned models were selected to participate in the combined ensemble predictions.
The ensemble handles videos directly as input, performing general preprocessing to separate the audio and video, extract frames of faces, and aggregate the scores from each model.

Each of these models focuses on different types of features. EfficientNet and XceptionNet mainly look at spatial features in video frames, MesoNet focuses on mesoscopic facial inconsistencies, and AASIST analyzes the audio to detect signs of spoofing. The idea behind using an ensemble is to take advantage of what each model does well and improve the overall detection performance.

The pipeline takes a video as input and first performs some general preprocessing. The video is split into audio and visual components, and frames are extracted at a fixed sampling rate. Faces are then detected in these frames using MTCNN. After this step, both the extracted faces and the audio are passed into separate preprocessing pipelines, since each model expects slightly different input formats and normalization. Once all the models produce their outputs, these scores are combined using an ensemble method to generate the final prediction.

\subsection{Data Preprocessing}

\subsubsection{Pipeline A: EfficientNet-B1 \& XceptionNet}

    Both EfficientNet and XceptionNet share the same preprocessing pipeline. When a video is fed into the ensemble, the ensemble initializes the Multi-task Cascaded Convolutional Neural Networks(MTCNN)\cite{mtcnn_docs}. The ensemble adopts the OpenCV library to extract every n frames from the video where sample rate = n. Then, the ensemble passes those frames extracted into MTCNN, which reports the coordinates of faces in those frames. The ensemble then manually extracts the faces from those frames using those coordinates.

    During the video frame extraction phase, OpenCV converts those frames from BGR to RGB, which is the standard input format for EfficientNet and XceptionNet.
    
    During face processing, EfficientNet first applies ImageNet-specific normalization by adjusting input video frames' pixel values. Specifically, the model subtracts the mean vector [0.485, 0.456, 0.406] from each pixel’s RGB values, followed by division by the standard deviation vector [0.229, 0.224, 0.225].
    
    After normalization is applied, EfficientNet uses the Albumentations library to squish or stretch every face such that it becomes 240 x 240 pixels.
    
    On the other hand, XceptionNet processes frames by converting them into RGB format, resizing them to 299 x 299 pixels, transforming them into a tensor, and normalizing them using a mean and standard deviation of 0.5 for each channel. 

\subsubsection{Pipeline B: MesoNet}

Although the MesoNet pipeline uses the same MTCNN preprocessing as Pipeline A, the model was integrated in a unique way. MesoNet was developed in 2018, but has not been regularly maintained in recent years. As such, to maximize replicating the original model at its time of publication, a separate Python virtual environment was generated to host legacy versions of MesoNet's dependencies (with the exception of the original Viola-Jones face detector, which has been replaced by the ensemble's MTCNN frame extractions). As such, the pipeline also serves to facilitate communication between the ensemble virtual environment and the MesoNet virtual environment using HTTP requests.

The MesoNet model shares similar preprocessing with Pipeline A, where the ensemble sends the extracted faces processed by MTCNN\cite{zhang2016mtcnn} to the MesoNet pipeline. The pipeline then performs some additional preprocessing to meet MesoNet specifications. First, the pipeline moves the PyTorch tensors from the GPU to the CPU, then converts the faces from its PyTorch tensor structure to a NumPy array. This is simply to send data designed for a PyTorch model to a Keras model, which MesoNet was implemented with. The pipeline then uses the OpenCV library\cite{bradski2000opencv} to resize the dimensions of the faces to a resolution of 256x256. If the pixel colors have not been normalized, then the pipeline will divide all pixel values by 255.0. The faces are then saved locally as a NumPy npy file, before the file path is sent with an HTTP request for the MesoNet model to find and load.

\subsubsection{Pipeline C: AASIST}
Unlike the video-based models, AASIST operates directly on raw audio waveforms rather than image data.

When a video is passed into the ensemble, the audio is first extracted using an external tool. The extracted audio is then converted to mono if necessary and resampled to a consistent format (typically 16 kHz). To ensure consistent input dimensions, the waveform is padded or truncated to a fixed length before being passed into the model \cite{aasist_repo}.

After preprocessing, the audio is fed into the AASIST model, which outputs logits corresponding to the bonafide and spoof classes. These logits are passed through a softmax function to produce a probability score, and the spoof probability is used as the final audio score.

This score is then passed into the ensemble and combined with the predictions from the visual models. Since AASIST focuses on detecting audio-based artifacts, it complements the video-based models by providing an additional modality for identifying deepfakes.

\subsection{Ensemble Logic}

The main reason for using an ensemble in this work was that deepfakes can contain different kinds of artifacts, and no single model is strong at detecting all of them. Some manipulations are easier to detect in the video frames, such as unnatural facial textures, blending errors, or inconsistencies around the mouth and eyes. Other manipulations are easier to detect in the audio, such as synthesized speech patterns or voice conversion artifacts. Because of this, combining multiple models is preferable to depending on only one \cite{dietterich2000ensemble}. In our case, the ensemble brings together AASIST for audio analysis and EfficientNet, MesoNet, and XceptionNet for visual analysis. Each model contributes a different perspective, so the overall system is intended to be more robust than any individual model by itself. 

The pipeline first separates the video into audio and visual information. The audio is passed to AASIST, while the visual part is processed frame-by-frame after face detection using MTCNN. The video models then produce their own fake-probability scores, and these outputs are collected for the ensemble step. One important design decision here is that the system keeps the individual scores instead of immediately collapsing them into a hard real/fake label. This is useful because a score contains more information than a binary output. For example, a score of 0.51 and a score of 0.99 would both count as “fake” after thresholding, but they represent very different levels of model confidence. Keeping the raw scores allows the fusion method to use that extra information. 

One fusion method used in the system is mean fusion. In this method, the available scores are averaged to produce one final confidence value. The logic behind this choice is mainly simplicity and stability. Mean fusion is easy to implement, easy to interpret, and gives every contributing model equal influence. This makes it a good baseline, especially early in development when it is still important to verify that each individual model is working properly. Another reason this choice makes sense is that averaging tends to reduce the effect of one unusually high or low score, so it can smooth out noise from individual models. In the current implementation, missing scores are ignored and only valid scores are averaged, which makes the system more practical since audio extraction or one of the visual models may occasionally fail. 

However, mean fusion also has limitations. Its main weakness is that it assumes all models are equally reliable, which is usually not true in practice. For example, one model may perform much better on certain datasets, while another may be less stable or more likely to fail. In that situation, treating every model equally can hold back performance. This is one of the reasons why a more advanced fusion strategy, stacking, was also introduced.

In stacking, the outputs of the base models are not simply averaged. Instead, they are passed into a separate meta-model that learns how to combine them \cite{wolpert1992stacked}. In this work, the stacking model takes four features in a fixed order: AASIST, EfficientNet, MesoNet, and XceptionNet scores. The reason for using stacking is that it allows the ensemble to learn relationships between model outputs instead of relying on a hand-designed rule. For instance, the meta-model can learn that one model should be trusted more when another model is uncertain, or that a certain combination of scores is strongly associated with fake videos \cite{geron2019hands}. This makes stacking more flexible than plain averaging.


Another key design choice in the stacking logic is how missing predictions are handled. If one model does not produce a score, the system replaces that missing value with 0.5. The reason for choosing 0.5 is that it acts like a neutral fallback. Since the scores are interpreted on a fake-probability scale, 0.5 means the model is maximally uncertain rather than strongly voting real or fake. This is a reasonable compromise because it prevents one failed model from forcing the ensemble too strongly in either direction. It also lets the stacking model continue working even when one branch of the pipeline breaks. 

The final ensemble output is interpreted using a threshold of 0.5, where values below 0.5 are classified as real and values at or above 0.5 are classified as fake. The logic behind this is consistency. Since the base models are already producing score-like outputs on a similar scale, using the same threshold across the system keeps the decision rule simple and easy to understand. It also makes the outputs easier to compare across models and across experiments. In practice, this means that a video is only classified as real when both the audio and visual components appear consistent with real data. If either modality produces a strong indication of manipulation, the final prediction is more likely to be classified as fake.

The system also includes a fallback from stacking back to mean fusion. This was a practical design choice. Stacking is more powerful, but it depends on an additional saved model file and can fail if the file is missing, corrupted, or incompatible. Rather than letting the whole pipeline crash, the code automatically switches back to mean fusion. The logic here is that a simpler fusion method is better than no output at all. This makes the pipeline more reliable and easier to diagnose during experimentation. 

Besides mean fusion and stacking, there are several other ensemble strategies that are relevant to this work. One of them is majority voting. In majority voting, each model first converts its score into a hard label, real or fake, and then the final decision is made based on whichever label gets the most votes. The advantage of majority voting is that it is simple and not overly affected by one extreme score. It can also work well when all models are reasonably calibrated and have similar importance. The downside is that it throws away confidence information. A weak fake prediction and a very strong fake prediction both count as only one vote, which can make the method less sensitive than score-based fusion. 

Another option is weighted average fusion. This is similar to mean fusion, except different models are given different importance weights. The reasoning behind this approach is that not all models perform equally well, so stronger models should influence the final result more. For example, if AASIST consistently performs better than one of the video models, its score could be given a larger weight. This can improve performance compared to a plain mean, but it also introduces another design problem: deciding what the weights should be. Those weights either need to be tuned manually or learned from data, and poor weight choices can hurt performance. 

A related approach is weighted voting, where models still cast binary votes, but some votes count more than others. This keeps the interpretability of voting while still reflecting differences in model quality. The limitation is similar to majority voting: once scores are converted into hard labels, much useful confidence information is lost. 

Overall, the progression of these options makes sense. Mean fusion is a strong starting point because it is simple, transparent, and stable. Majority voting is also simple, but it loses too much information from the raw scores. Weighted average and weighted voting try to fix the equal-importance problem, but they require extra tuning. Stacking goes one step further by learning the fusion rule directly from data, which is why it is the most flexible option out of the group \cite{wolpert1992stacked}. In our implementation, this made stacking the most advanced ensemble choice, while mean fusion served as the reliable fallback and baseline for comparison.

\section{EXPERIMENTS AND PROTOCOL}

\subsection{Datasets}

To determine whether the ensemble could make accurate predictions across multiple AI-generation methods, a variety of different datasets were sampled and used in the training, validating, and testing of the ensemble.

The models that relied on analyzing facial features were trained on videos from the AIGVDBench \cite{ma2026aigvdbench} and the FaceForensics++ dataset \cite{rossler2019faceforensics}. AIGVDBench contains 422 thousand general AI-generated videos, with content such as objects, landscapes, animals, people, or more. These videos were generated by one of 31 distinct models, with some models that were developed as recently as early 2025, and use methods involving text-to-video, image-to-video, and video-to-video. Although the generation models and methods are diverse, the dataset did not have an audio component with which to train AASIST. The AIGVDBench also did not focus exclusively on faces, hence only the subset of videos containing detectable faces was used to train early weights for the face-analyzing models, and some of these weights were carried over for further training using FaceForensics++.

FaceForensics++ was compiled in late 2019 consisting of videos generated using one of four different methods: Face2Face, FaceSwap, DeepFakes, and NeuralTextures. These methods focused particularly on clear, AI-manipulated faces, hence videos from FaceForensics++ were used to train the models that specialized in analyzing facial features. 700 real and 700 fake videos were used to train the models, with 150 real and 150 fake videos used to validate the individual model weights.

The audio-based model in the ensemble was trained using the ASVspoof 2019 Logical Access (LA) dataset \cite{todisco2019asvspoof}, which is a commonly used benchmark for detecting fake or manipulated speech. The dataset contains around 122,302 audio samples (about 7.5 GB total) and is split into three main parts: 25,381 samples for training, 24,987 for validation, and 71,934 for testing. Notably, the ASVspoof split differs from the typical 70/15/15 or 80/10/10 used by most datasets. Instead, it follows a 20/20/60 split. This means the test set is much larger than usual. The reason for this is to better evaluate how well models generalize to new, unseen types of spoofing attacks. Since the test set includes a wider variety of spoofing methods (some not seen during training), it pushes the model to learn more general patterns instead of just memorizing specific examples. Also, the split is predefined using protocol files, so there is no randomness involved, which helps keep results consistent across different experiments. The spoofed audio is created using different techniques like text-to-speech (TTS) and voice conversion (VC), which adds diversity and makes the training more realistic. \cite{todisco2019asvspoof}.

As the ensemble unified several models, each specialized in analyzing different video components, an equally diverse dataset was required. This dataset must not only be a suitable input to all models, but also reduce any bias towards a specific generation technique. Hence the FakeAVCeleb dataset \cite{khalid2021fakeavceleb} became a viable candidate to test the ensemble, as the dataset explores the permutations of image and audio generation. FakeAVCeleb was compiled in 2021 with videos comprised of 5 different ethnic backgrounds, containing 500 videos with real audio and video, 500 videos with fake audio and real video, 9000 videos with fake video and real audio, and 10000 videos with fake audio and fake video. This amounts to a total of 500 real videos and 19500 deepfake videos, which were divided in half to serve as a validation and test set for the ensemble.\\
In the following sections, let $R$ and $F$ denote real and fake content, respectively. For example, $RvFa$ represents a composite sample consisting of real video ($v$) and fake audio ($a$).
Due to the significant class imbalance in the original FakeAVCeleb dataset, which is biased toward manipulated content, we constructed a balanced test set by randomly sampling an equal number of videos from both the real(RvRa) and fake(RvFa, FvRa, FvFa) categories.
\subsection{Training Protocol}

The hyperparameters used for face detection and for each model are summarized in Tables \ref{table:mtcnn_config}--\ref{table:xception_hyperparameters}.

\begin{table}[H]
    \centering
    \caption{MTCNN Hyperparameters}
    \label{table:mtcnn_config}
    \begin{tabular}{ll}
        \hline
        \textbf{Hyperparameter} & \textbf{Value} \\
        \hline
        Stride & 1 \\
        Resize & 1.0 \\
        Margin & 200 \\
        Minimum Face Size & 100 \\
        Thresholds & [0.6, 0.7, 0.7] \\
        Scale Factor & 0.7 \\
        Post-Process & True \\
        Select Largest & True \\
        Keep All & True \\
        \hline
    \end{tabular}
\end{table}

\begin{table}[H]
    \centering
    \caption{EfficientNet Hyperparameters}
    \label{table:training_hyperparameters}
    \begin{tabular}{ll}
        \hline
        \textbf{Hyperparameter} & \textbf{Value} \\
        \hline
        \multicolumn{2}{c}{\textit{Architecture \& Data}} \\
        \hline
        Model Architecture & EfficientNet-B1 \\
        Input Resolution & $240 \times 240$ \\
        Data Augmentation & Heavy Augmentation \\
        \hline
        \multicolumn{2}{c}{\textit{Optimization}} \\
        \hline
        Optimizer & AdamW \\
        Loss Function & Cross-Entropy \\
        Base Learning Rate & $8 \times 10^{-4}$ \\
        Weight Decay & $1 \times 10^{-3}$ \\
        \hline
        \multicolumn{2}{c}{\textit{Training Schedule}} \\
        \hline
        Batch Size & 32 \\
        Epochs & 5 \\
        Framework & PyTorch \\
        Device & CUDA (GPU) \\
        \hline
    \end{tabular}
\end{table}

\begin{table}[H]
    \centering
    \caption{AASIST Hyperparameters}
    \label{table:aasist_hyperparameters}
    \begin{tabular}{ll}
        \hline
        \textbf{Hyperparameter} & \textbf{Value} \\
        \hline
        \multicolumn{2}{c}{\textit{Architecture \& Data}} \\
        \hline
        Model Architecture & AASIST (Audio Anti-Spoofing) \\
        Input Type & 16 kHz Mono Audio \\
        Feature Representation & Raw waveform (learned features) \\
        \hline
        \multicolumn{2}{c}{\textit{Optimization}} \\
        \hline
        Optimizer & Adam \\
        Loss Function & Binary Cross-Entropy \\
        Learning Rate & $1 \times 10^{-4}$ \\
        \hline
        \multicolumn{2}{c}{\textit{Training Schedule}} \\
        \hline
        Batch Size & 24 \\
        Number of Epochs & 10 \\
        Device & CUDA (GPU) \\
        \hline
    \end{tabular}
\end{table}

\begin{table}[H]
    \centering
    \caption{MesoNet Hyperparameters}
    \label{table:mesonet_hyperparameters}
    \begin{tabular}{ll}
        \hline
        \textbf{Hyperparameter} & \textbf{Value} \\
        \hline
        \multicolumn{2}{c}{\textit{Architecture \& Data}} \\
        \hline
        Model Architecture & Meso4 \\
        Input Resolution & $256 \times 256$ \\
        Input Type & Face Crops (RGB) \\
        \hline
        \multicolumn{2}{c}{\textit{Optimization}} \\
        \hline
        Optimizer & Adam \\
        Loss Function & Binary Cross-Entropy \\
        Learning Rate & $1 \times 10^{-3}$ (pretrained) \\
        \hline
        \multicolumn{2}{c}{\textit{Training Schedule}} \\
        \hline
        Batch Size & 32 (pretrained) \\
        Number of Epochs & 7+7 (AIGVDBench + FaceForensics++) \\
        Weights & meso4\_weight\_refit\_val\_acc.h5 \\
        Framework & Keras 2.1.5 (served via FastAPI) \\
        \hline
    \end{tabular}
\end{table}

\begin{table}[H]
    \centering
    \caption{XceptionNet Hyperparameters}
    \label{table:xception_hyperparameters}
    \begin{tabular}{ll}
        \hline
        \textbf{Hyperparameter} & \textbf{Value} \\
        \hline
        \multicolumn{2}{c}{\textit{Architecture \& Data}} \\
        \hline
        Model Architecture & XceptionNet \\
        Input Resolution & $299 \times 299$ \\
        Input Type & Face Crops (RGB) \\
        Pretraining & ImageNet \\
        \hline
        \multicolumn{2}{c}{\textit{Optimization}} \\
        \hline
        Optimizer & Adam \\
        Loss Function & Binary Cross-Entropy \\
        Learning Rate & $2 \times 10^{-4}$ (fine-tuning) \\
        \hline
        \multicolumn{2}{c}{\textit{Training Schedule}} \\
        \hline
        Batch Size & 32 \\
        Number of Epochs & 10--20 \\
        Fine-Tuning & Fully Connected Layers + Partial Backbone \\
        Framework & PyTorch \\
        \hline
    \end{tabular}
\end{table}
\subsection{Evaluation Metrics}
To evaluate the performance of the individual models and the ensemble, we primarily use accuracy as the evaluation metric. For each video in the dataset, each individual model is evaluated against the same ground-truth label of the original video. The ensemble is evaluated against the overall ground-truth label, where a video is considered real only if both its video and audio components are real, and fake otherwise. Finally, accuracy is defined as the proportion of correctly classified samples over the total number of samples.

\section{RESULTS}
\subsection{AASIST}
As a standalone model, AASIST performed very well on the ASVspoof2019 Logical Access (LA) dataset, achieving a best development Equal Error Rate (EER) of 0.94\% and an accuracy of 99.16\% when using pretrained weights provided with the model. It was especially good at detecting synthetic or voice-cloned audio by picking up on subtle differences in the sound patterns. However, one limitation is that it cannot detect deepfakes that only manipulate the video and not the audio. Its performance can also drop if the audio preprocessing is not done properly, such as when the sampling rate is incorrect or there is excessive noise. 
\subsection{EfficientNet}
When EfficientNet-B1 was trained on the AIGVDBench dataset, the final accuracy on FakeAVCeleb validation set was 53\%.
When EfficientNet-B1 was trained on the FaceForensics++ dataset, the final accuracy on FakeAVCeleb validation set was 72\%.
These are facial accuracy, which calculates number of correctly predicted faces divided by number of total faces.
Once EfficientNet is integrated into the ensemble, we use video accuracy, similar to facial accuracy, the video accuracy is calculated as the number of correctly predicted videos/the number of total videos.
By tuning the threshold, EfficientNet's video accuracy can scale from 30\% to 61\%.

Based on the result, EfficientNet struggles to generalize when being trained on only one dataset. Because EfficientNet processes faces one by one to make individual predictions, it cannot capture temporal flickering or subtle deepfake traces that only appear during continuous video playing.

\subsection{XceptionNet}
The GitHub repository of the XceptionNet model used in this ensemble provides two pretrained weights, \texttt{FF++\_c23.pth} and \texttt{FF++\_c40.pth}, which were pretrained on the FaceForensics++ dataset. In particular, \texttt{FF++\_c23.pth} was chosen. Before joining the ensemble, this model was trained, validated, and tested on the videos provided by the AIGVDBench dataset and the FaceForensics++ dataset. The results are shown in Table \ref{tab:results}.

\begin{table}[h]
\centering
\begin{tabular}{|c|c|c|}
\hline
Dataset & Validation accuracy & Test accuracy \\
\hline
AIGVDBench & 95.07\% & 95.05\% \\
FaceForensics++ & 93.64\% & 90.16\% \\
\hline
\end{tabular}
\caption{Dataset accuracy comparison}
\label{tab:results}
\end{table}

Moreover, this model was also tested on the FakeAVCeleb dataset and achieved an accuracy of 60.40\%. This suggests that XceptionNet performs well at detecting deepfakes in datasets that it was trained on. However, the model struggles when tested on a more general dataset such as FakeAVCeleb, which suggests that the model learns dataset-specific artifacts rather than generalizable deepfake features. 

\subsection{MesoNet}
Although the original model remains unchanged, new weights were trained using videos provided by AIGVDBench\cite{ma2026aigvdbench}, which were then reused and trained with videos provided by FaceForensics++\cite{rossler2019faceforensics}. For both dataset trainings, maximization of the validation accuracy was used as a metric to determine a more valuable weight. Other weights were also trained that prioritized minimizing the validation loss between training epochs.

Initial testing was conducted using the original Meso4 DeepFake weight and pipeline provided by the MesoNet authors to evaluate how well the older weights and model may perform. The Meso4 architecture produced an accuracy of 52.083\% on videos from the AIGVDBench dataset, labeling 84.462\% of real videos correctly and 19.704\% of AI-generated videos correctly. This was not surprising, since the weights were produced 6 years prior and were specialized on DeepFake videos, not the current AIGVDBench. Using the original weight with FaceForensics++, MesoNet achieved a 61.65\% accuracy on DeepFake videos.

New weights were trained and validated with videos from the AIGVDBench dataset, through MTCNN \cite{zhang2016mtcnn} as the method to identify and extract faces. The weight produced through AIGVDBench achieved a validation accuracy of 85.06\% and served as the initial weight for training and validation using the FaceForensics++ dataset. With this, the final weight produced a validation accuracy of 90.86\% on FaceForensics++, and is used for testing within the ensemble.

The main strength of the model was its ability to make accurate predictions with significantly fewer parameters. The model required less computational power than most models, allowing faster training and predictions with fewer resources. However, the low number of parameters can also contribute to a significant weakness; the model would specialize only on certain AI-generation methods instead of identifying overall patterns between videos.

\subsection{Ensemble}
\subsubsection{FakeAVCeleb Validation Set result}
The FakeAVCeleb dataset has 21544 videos in total. The manipulation categories are as follows:
\begin{itemize}
    \item FakeVideo-FakeAudio(FvFa): 10835
    \item FakeVideo-RealAudio(FvRa): 9709
    \item RealVideo-FakeAudio(RvFa): 500
    \item RealVideo-RealAudio(RvRa): 500
\end{itemize}
The dataset is heavily skewed toward fake videos.
The whole test set contains half the number of videos from each of the categories above. 
\subsubsection{Ensemble method results}
\begin{itemize}

    \item \textbf{Weighted voting:}\\

    \begin{table}[H]
        \centering
        \caption{FvFa and RvFa and FvRa and RvRa}
        \label{table:performance_summary_best_config}
        \begin{tabular}{lc}
                \hline
                \textbf{Model} & \textbf{Accuracy} \\
                \hline
                EfficientNet & 0.500 \\
                MesoNet      & 0.696 \\
                XceptionNet  & 0.698 \\
                AASIST       & 0.494 \\
                \hline
                Ensemble     & 0.728 \\
                \hline
            \end{tabular}
        \footnotesize
    \end{table}

    \item \textbf{Majority voting:}\\
    
    \begin{table}[H]
        \centering
        \caption{FvFa and RvFa and FvRa and RvRa }
        \label{table:performance_summary_majority_voting}
            \begin{tabular}{lc}
                \hline
                \textbf{Model} & \textbf{Accuracy} \\
                \hline
                EfficientNet & 0.500 \\
                MesoNet      & 0.696 \\
                XceptionNet  & 0.698 \\
                AASIST       & 0.498 \\
                \hline
                Ensemble     & 0.712 \\
                \hline
            \end{tabular}
        \footnotesize
    \end{table}

    \item \textbf{Mean:}\\
    
    \begin{table}[H]
        \centering
        \caption{FvFa and RvFa and FvRa and RvRa }
        \label{table:performance_summary_mean}
        \begin{tabular}{lc}
            \hline
            \textbf{Model} & \textbf{Accuracy} \\
            \hline
            EfficientNet & 0.500 \\
            MesoNet      & 0.696 \\
            XceptionNet  & 0.698 \\
            AASIST       & 0.500 \\
            \hline
            Ensemble     & 0.616 \\
            \hline    
        \end{tabular}
        \footnotesize
    \end{table}

    \item \textbf{Stacking:}\\
    
    \begin{table}[H]
        \centering
        \caption{FvFa and RvFa and FvRa and RvRa }
        \label{table:performance_summary_stacking}
        \begin{tabular}{lc}
            \hline
            \textbf{Model} & \textbf{Accuracy} \\
            \hline
            EfficientNet & 0.500 \\
            MesoNet      & 0.696 \\
            XceptionNet  & 0.698 \\
            AASIST       & 0.494 \\
            \hline
            Ensemble     & 0.710 \\
            \hline
        \end{tabular}
        \footnotesize
    \end{table}

    \item \textbf{Weighted average:}\\
    
    \begin{table}[H]
        \centering
        \caption{FvFa and RvFa and FvRa and RvRa} 
        \label{table:performance_summary_weighted_average}
        \begin{tabular}{lc}
            \hline
            \textbf{Model} & \textbf{Accuracy} \\
            \hline
            EfficientNet & 0.532 \\
            MesoNet      & 0.602 \\
            XceptionNet  & 0.658 \\
            AASIST       & 0.490 \\
            \hline
            Ensemble     & 0.658 \\
            \hline
        \end{tabular}
        \footnotesize
    \end{table}

\end{itemize}
\subsubsection{Performance of the Best Ensemble Method Across Manipulation Categories}

\begin{itemize}
    \item \textbf{FvFa and RvRa:}\\
    \begin{table}[H]
        \centering
        \caption{Performance on Balanced FvFa RvRa Test Set}
        \label{table:balanced_fvfa_rvra}
        \begin{tabular}{lc}
            \hline
            \textbf{Model} & \textbf{Accuracy} \\
            \hline
            EfficientNet & 0.500 \\
            MesoNet      & 0.782 \\
            XceptionNet  & 0.806 \\
            AASIST       & 0.464 \\
            \hline
            Ensemble     & 0.848 \\
            \hline
        \end{tabular}
        \footnotesize
    \end{table}

    \item \textbf{FvRa and RvRa:}\\
    \begin{table}[H]
        \centering
        \caption{Performance on Balanced FvRa RvRa Test Set}
        \label{table:balanced_fvra_rvra}
        \begin{tabular}{lc}
            \hline
            \textbf{Model} & \textbf{Accuracy} \\
            \hline
            EfficientNet & 0.500 \\
            MesoNet      & 0.720 \\
            XceptionNet  & 0.744 \\
            AASIST       & 0.484 \\
            \hline
            Ensemble     & 0.774 \\
            \hline
        \end{tabular}
        \footnotesize
    \end{table}

    \item \textbf{RvFa and RvRa:}\\
    \begin{table}[H]
        \centering
        \caption{Performance on Balanced RvFa RvRa Test Set}
        \label{table:balanced_rvfa_rvra}
        \begin{tabular}{lc}
            \hline
            \textbf{Model} & \textbf{Accuracy} \\
            \hline
            EfficientNet & 0.499 \\
            MesoNet      & 0.515 \\
            XceptionNet  & 0.501 \\
            AASIST       & 0.532 \\
            \hline
            Ensemble     & 0.514 \\
            \hline
        \end{tabular}
        \footnotesize
    \end{table}

\end{itemize}

\section{DISCUSSION}
\subsection{Generalization issue}
By comparing AASIST accuracy during the validation phase, we conclude that AASIST does not generalize well. When half of the videos in a test contain real audio while the other half contain fake audio, the accuracy is around 50\%, indicating near-random performance. When all of the videos in a test set contain real audios, accuracy jumps to 90\%. AASIST collapses to predicting real on FakeAVCeleb audio, indicating a substantial domain gap between benchmark and in-the-wild speech. However, EfficientNet, MesoNet, and XceptionNet generalize relatively well, as the accuracy is roughly 70\% to 80\% regardless of how many fake videos are present in a test set. This indicates that the video-based models are more robust to changes in dataset composition.

It is also important to note that the dataset used in this work is heavily skewed toward fake samples. This imbalance can bias models toward predicting the majority class and may partially explain inconsistent behavior across different test configurations. As a result, accuracy alone may not fully reflect true model performance.

\subsection{Concrete votes reduce random noise}
Weighted voting and majority voting outperformed weighted average and mean. This suggests that forcing models to make concrete predictions can effectively reduce noise and result in a more reliable ensemble prediction.

Instead of allowing small fluctuations in confidence scores to influence the final result, voting-based methods emphasize agreement between models. This leads to more stable and reliable ensemble predictions.

However, the improvement from ensemble methods is relatively small compared to the strongest individual model. This suggests that the models may not be sufficiently diverse, or that some models share similar weaknesses. As a result, the ensemble is limited in its ability to significantly outperform its best performing components.

\subsection{The ensemble relies heavily on the video models}
In the validation phase, we have noticed AASIST does not generalize well to distinguish fake audio from real audio. This is shown again in Table 14, which exhibits a large decrease in ensemble accuracy compared to Tables 13 and 12. The fundamental change is that there is no video manipulation artifact in the test set of Table 14. However, the video models can contribute effectively to raise ensemble accuracy above their individual accuracy. 

This indicates that the ensemble relies heavily on the video-based models for its performance. When visual artifacts are present, the video models contribute effectively and can even push the ensemble accuracy beyond their individual performance.

Additionally, not all models contribute equally to the ensemble. In several configurations, AASIST and EfficientNet perform close to random, suggesting that the overall improvement is primarily driven by stronger models such as XceptionNet and MesoNet. This creates an imbalance in the multimodal setup, where audio information is not being fully utilized.

\subsection{Individual models must be sensitive to catch fake videos}
Because the video models were not trained with the same manipulation technique, the video decision threshold present in each ensemble method is around 0.2--0.4. This forces the video models to be highly sensitive to produce a good result. 

This suggests that, in practice, the video models are encouraged to classify a video as fake even when only subtle or partial artifacts are present. Such sensitivity is important because deepfake techniques vary widely, and models may encounter manipulation patterns that were not seen during training. Being more sensitive allows the ensemble to better capture these unseen artifacts, although it may also increase the risk of false positives.

\subsection{Summary}
Overall, the results show that the performance of the system is strongly influenced by both model generalization and the way predictions are combined. While the ensemble improves consistency compared to individual models, its effectiveness is limited by the weakest component, particularly the audio model.

The results also demonstrate that different fusion strategies lead to different trade-offs, where voting-based methods provide more stable predictions by reducing noise, while score-based methods are more sensitive to unreliable outputs. However, the improvement from the ensemble is relatively modest, suggesting that not all models contribute equally and that some share similar weaknesses.

In addition, the system is largely driven by the video-based models, as they are able to generalize more effectively and detect a wider range of manipulation artifacts. However, this creates an imbalance in the multimodal setup, where the audio modality remains underused. Combined with the effects of dataset imbalance, this highlights the importance of both improving individual model performance and designing more effective fusion strategies.

Finally, the need for high sensitivity in video models highlights the challenge of detecting unseen manipulation techniques, reinforcing the importance of designing models that can generalize beyond the data they were trained on.

\section{LIMITATIONS AND FUTURE WORK}
One important limitation of deepfake detection models, including AASIST and the video-based models used in this work, is that they do not always generalize well across different datasets. These models can perform well when they are trained and tested on the same dataset, but their performance often drops when they are tested on new or unseen data. This mainly happens because of the domain gap, where different datasets have variations in factors such as lighting, resolution, compression, and even the types of deepfake techniques used \cite{yin2021dabn}. 

As mentioned in previous work, many detection models end up learning dataset-specific patterns instead of more general features, which makes them less reliable in real-world situations \cite{yin2021dabn}. This becomes even more of a problem as new deepfake generation methods continue to improve, since models trained on older data may not be able to detect newer types of fakes effectively. 

One method that tries to deal with this issue is Domain Adaptive Batch Normalization (DABN). Normally, batch normalization layers use statistics like the mean and variance that were calculated during training. The problem is that if the test data looks different from the training data, those statistics might not match anymore, which can hurt performance. DABN fixes this by updating those statistics during testing, using values from the test data instead. This helps the model adjust to the new data distribution without needing to retrain the whole network \cite{yin2021dabn}. 

By doing this, DABN helps reduce the effect of domain shift and makes the model more robust when working with different datasets. It is also a fairly flexible approach since it can be applied to different models without major changes. 

Another limitation of the current work is that the contribution of each individual model within the ensemble was not explicitly analyzed. While the ensemble improves overall performance, it is not always clear how much each model contributes to the final prediction. Conducting ablation studies where models are systematically removed or evaluated in isolation would provide better insight into the importance of each component and help guide future improvements to the ensemble design.

For future work, improving generalization remains a significant challenge. This could include using more diverse datasets during training, applying better domain adaptation techniques, or combining audio and video information more effectively in a multimodal setup. Even though methods like DABN help, making models that generalize well to completely new types of deepfakes is still an open problem.

\section{CONCLUSION}
This work developed a deepfake detection pipeline that combines both audio and visual information using an ensemble of four models: EfficientNet-B1, XceptionNet, MesoNet, and AASIST. The system processes video by separating it into frames and audio, applying model-specific preprocessing, and then combining the resulting predictions through different ensemble methods.
The results of this work show that while individual models perform well on the datasets they are trained on, their performance drops when applied to more diverse data. This highlights a key limitation of single-model approaches in deepfake detection. By combining multiple models, the ensemble was able to produce more consistent predictions across different types of deepfakes, improving overall robustness.
Mean fusion served as a stable and reliable baseline, while stacking provided a more flexible way to combine model outputs when the additional model was available. However, the overall accuracy of approximately 70\% suggests that there are still challenges in achieving strong generalization across datasets.
Overall, this work suggests that using both audio and visual information together could provide more information than relying on a single model. At the same time, it shows that domain gap and model generalization remain important areas for further improvement.

\section{CODE AVAILABILITY}

The implementation of the ensemble pipeline, including model wrappers, preprocessing code, and the FastAPI backend, is publicly available at: \url{https://github.com/gdgutm/AI-Video-Detection}.

\section{ACKNOWLEDGMENTS}

Thank you to the University of Toronto for the academic environment and resources necessary for this research. During the preparation of this work, the authors used Claude (Anthropic) for language editing, proofreading, and LaTeX formatting assistance. The authors reviewed and edited all content and take full responsibility for the final manuscript.

\nocite{*}
\printbibliography[title={REFERENCES}, heading=bibintoc]

\end{document}